\documentclass{article}

\usepackage{arxiv}

\usepackage[utf8]{inputenc} 
\usepackage[T1]{fontenc}    
\usepackage{hyperref}       
\usepackage{url}            
\usepackage{booktabs}       
\usepackage{amsfonts}       
\usepackage{nicefrac}       
\usepackage{microtype}      
\usepackage{lipsum}		
\usepackage{graphicx}
\usepackage{natbib}
\usepackage{doi}
\usepackage{amsmath}
\usepackage{epsfig}
\usepackage{appendix}
\usepackage{placeins} 
\usepackage{float}    

\title{Comparative Analysis of Different Efficient Fine Tuning Methods of Large Language Models (LLMs) in Low-Resource Setting}

\author{ \hspace{1mm}Krishna Prasad Varadarajan Srinivasan \\
	Georgia Institute of Technology \\
	\texttt{kvarada@gatech.edu} \\
	\And
	\hspace{1mm}Prasanth Gumpena \\
	Georgia Institute of Technology\\
	\texttt{pgumpena3@gatech.edu} \\
        \And
        \hspace{1mm}Madhusudhana Yattapu \\
	Georgia Institute of Technology\\
	\texttt{myattapu3@gatech.edu} \\
        \And
        \hspace{1mm}Vishal H. Brahmbhatt \\
	Georgia Institute of Technology\\
	\texttt{vbrahmbhatt3@gatech.edu} \\
}

\date{}

\hypersetup{
pdftitle={A template for the arxiv style},
pdfsubject={q-bio.NC, q-bio.QM},
pdfauthor={David S.~Hippocampus, Elias D.~Striatum},
pdfkeywords={First keyword, Second keyword, More},
}

\begin{document}
\maketitle

\begin{abstract}

 In the domain of large language models (LLMs), \cite{mosbach2023fewshot} showed that few-shot full-model fine-tuning -- namely Vanilla Fine Tuning (FT) and Pattern-Based Fine Tuning (PBFT) --, and In-Context Learning (ICL) generalize similarly on Out-Of-Domain (OOD) datasets, but vary in terms of task adaptation. However, they both pose challenges, especially in term of memory requirements. In this paper, we further try to push the understanding of different fine-tuning strategies for LLM and aim to bring a myriad of these on the same pedestal for an elaborate comparison with full-model fine-tuning on two diverse datasets. To that end, we conducted a series of experiments, beginning with state-of-the-art methods like vanilla fine-tuning and Pattern-Based Fine-Tuning (PBFT) on pre-trained models across two datasets, COLA and MNLI. We then investigate adaptive fine-tuning and the efficiency of LoRA adapters in a few-shot setting. Finally, we also compare an alternative approach that has gained recent popularity -- context distillation -- with the vanilla FT and PBFT with and without few-shot setup. \\
 
   \indent Our findings suggest that these alternative strategies that we explored can exhibit out-of-domain generalization comparable to that of vanilla FT and PBFT.  PBFT under-performs Vanilla FT on out-of-domain (OOD) data, emphasizing the need for effective prompts. Further, our adaptive-fine tuning and LoRA experiments perform comparable or slightly worse than the standard fine-tunings as anticipated, since standard fine-tunings involve tuning the entire model. Finally, our context distillation experiments out-perform the standard fine-tuning methods. These findings underscore that eventually the choice of an appropriate fine-tuning method depends on the available resources (memory, compute, data) and task adaptability. \\
\end{abstract}


\section{Introduction}
\noindent The goal of our work is to evaluate and compare the performance of a pre-trained large language model on sequence classification tasks. We aimed to do this by employing -- 1) different fine tuning (FT) methods, 2) applying Low-Rank Adaptation - LoRA (\cite{hu2021lora}) adaptors with few-shot learning, and 3) performing context-distillation both with and without few-shot learning setting. We aim to understand the efficacy of alternative fine-tuning methods on a pre-trained large language model’s performance in sequence classification tasks using 2 datasets, namely \href{https://paperswithcode.com/dataset/multinli}{MNLI} (\cite{N18-1101}) and \href{https://nyu-mll.github.io/CoLA/}{COLA} (\cite{warstadt2018neural}), which are further explained in Section \ref{Datasets}. We explored alternate ways of efficiently fine-tuning the model and compared them with the baseline methods (vanilla and pattern-based fine-tuning) for Open Pre-trained Transformer (OPT) (\cite{zhang2022opt}) model's performance on the text sequence classification task using both in-domain and out of domain accuracies. We try to keep the training process, experiments, and hyper-parameters similar across various experiments, wherever possible, for a fair comparison. 

Currently large language models (LLMs) are pre-dominantly used by leveraging In-Context Learning - ICL (\cite{brown2020language}), whereby during the inference time, the model learns to answer follow-up questions from a series of prompts. This approach requires significant inference time memory and compute. In recent times, bunch of alternate methods have been proposed and explored to augment the issues faced with ICL. We explore and analyze a number of these methods in our work presented here.

Our work is relevant to anyone leveraging large language models for tasks like chat bots and code completion. If successful, our methods could improve the efficiency and performance of these models, particularly in tasks requiring the ingestion of large sequences of dialogue, code, or text.

We expand on all of these methods in section \ref{Approach} of this paper. We also further expand on the 2 datasets that we have performed our experiments on, in the section \ref{Datasets}.

\subsection{Datasets} \label{Datasets}
\subsubsection{MNLI}

Multi-Genre Natural Language Inference - MNLI (\cite{N18-1101}) is a crowd-sourced collection of sentence pairs with textual entailment annotations. This is one of the largest corpora available for natural language inference (NLI) with ten distinct genres of written and spoken English. This paper used a subset of 261,802 samples accessed via GLUE (\cite{wang2019glue}) for training and entire dataset for in-domain-performance. Note that the labels were binarized to only include entailment and contradiction. The most important aspects of the dataset are suitability for NLI task and variety of samples belonging to different genres. An example of premise, hypothesis, and label are - \\
\noindent \textbf{Premise:} How do you know? All this is their information again. \\
\textbf{Hypothesis:} This information belongs to them. \\
\textbf{Label:} 0 (\textit{entailment}) \\

 HANS dataset (\cite{mccoy2019right}) with 30,000 samples, was used for out-of-domain performance. It was chosen to evaluate performance when training on the MNLI dataset using different methods. This dataset includes examples that challenge conventional MNLI patterns by failing three syntactic heuristics - lexical overlap, subsequence, and constituent. These examples test how well models handle misleading or incorrect heuristic cues. \\
\noindent \textbf{Premise:} The senators supported the actor. \\
          \textbf{Hypothesis:} The actor supported the senators. \\
          \textbf{Label:} 1 (\textit{positive})

\subsubsection{COLA}

\noindent The Corpus of Linguistic Acceptability (COLA) (\cite{warstadt2018neural}), developed by researchers at New York University (NYU) and Meta's AI research lab, encompasses 10,657 English sentences from 23 linguistic publications. It provides annotations on sentence grammaticality, with 9,594 sentences for training and 1,063 for testing. These annotations are provided by the authors of the sources. COLA aims to uniquely explore neural networks' capacity to understand grammatical nuances.

The dataset uses labels for grammatical acceptability: \textbf{0} - unacceptable, \textbf{1} - acceptable. For instance: \
\textbf{Sentence:} This paper was written in Overleaf. \
\textbf{Label:} 1 (\textit{acceptable}) \

Among the 23 sources, 17 are deemed in-domain, while the remaining 6 are out-of-domain (OOD). Access to the in-domain data was via the GLUE framework (\cite{wang2019glue}), and the OOD data from (\cite{meta2023efficientllm}) GitHub repository, allowing for consistent analysis.

\section{Approach} \label{Approach}

\noindent We aimed to enhance sequence classification performance of large language models. Using fine-tuning methods like vanilla and Pattern-Based Fine Tuning (PBFT), we established baselines on Facebook’s OPT 125M and OPT 350M models in a few-shot setting. Limited by time and resources, we focused solely on these models. Our baseline experiments, which replicated (\cite{mosbach2023fewshot}), results for Vanilla and PBFT methods, used both datasets mentioned in Section \ref{Datasets}.

We broadened our experiments to include Context Distillation, as per Anthropic’s discussion in \cite{askell2021general}. We experimented this fine-tuning in both few-shot and standard scenarios, with performance being comparable. We also tested Parameter-Efficient Fine-Tuning (PEFT) enhanced with Low-Rank Adaptation - LoRA (\cite{hu2021lora}) and Adaptive Fine-Tuning. Our findings indicate that while all methods perform similarly, they can be unstable and under-perform due to training instability. Performance generally improves with model size, and context distillation often excels in generalization.  A comprehensive analysis is provided in Section \ref{Experiments and Results}.

Our approach, a unique mix of traditional and new methods, shows potential. While similar methods may have been individually explored, we focus on their benefits when applied separately and the synergy when few-shot learning is combined with LoRA adapter or context distillation. We aim to thoroughly analyze their performance in sequence classification tasks, hoping our exploration offers valuable insights to the field.

We foresaw challenges with computational resources and time, leading us to focus on the OPT 125M and OPT 350M models. We also expected performance differences between in-domain and out-of-domain accuracies. A main challenge was ensuring fair comparison across methods, especially as context distillation had significantly more training examples on the full set than in a few-shot setting.

\subsection{Few-Shot Learning} \label{Few-Shot Learning}

\noindent For our experiments, we used few-shot learning to design models that learn from a small number of examples, using knowledge from related tasks. We conducted experiments with varying numbers of examples (N = 2, 16, 32, 64, and 128), fine-tuning and evaluating models in each setting. This process, repeated across all settings, allowed us to observe the impact of example quantity on model performance. To ensure robust results, we performed 10 runs per model size per N, accounting for data seed variability.

\subsection{Infrastructure and Replicating the Codebase}

\noindent The experiments were conducted using Google Colab Pro, equipped with Nvidia Tesla T4 and L4 GPUs.  The entire codebase for our work is hosted on GitHub and can be accessed at \url{https://github.com/iamvarada/llm_finetuning} \footnote{Please contact the authors for any access issues}. Researchers interested in replicating or extending our work are encouraged to clone or fork this repository. To ensure consistency and ease of setup, all necessary Python dependencies are listed in the \textit{requirements.txt} file within the repository.  

\section{Fine-Tuning Methods}

\subsection{Vanilla Fine-Tuning}
\noindent Vanilla fine-tuning was performed on two models, OPT 125M and OPT 350M, instantiated as SequenceClassification with dual classification heads for binary labels. The models were directly fed with the ‘premise’ and ‘hypothesis’, and respective labels for MNLI and COLA datasets, without any prompts. The training loop, focused on few-shot learning as explained in Section \ref{Few-Shot Learning}, 

\subsection{Pattern-Based Fine-Tuning}
\noindent Pattern-Based Fine-Tuning (PBFT) is an advanced approach that uses language patterns to guide fine-tuning. Unlike vanilla fine-tuning, PBFT uses specific prompts or patterns to guide model predictions. We use GPT-3 derived patterns to transform the ‘premise’ and ‘hypothesis’ into a more learnable format. The model is then fine-tuned on these transformed inputs. PBFT, particularly with GPT-3 patterns, leverages inductive biases in prompts for effective learning and better task performance, especially useful in few-shot learning scenarios with limited data.

\subsection{Adaptive Fine-Tuning}
\noindent Adaptive fine-tuning enables nuanced training using two techniques: Freezing Layers and Dynamic Learning Rate. The initial layers of the OPT 125M model, which hold general-purpose knowledge, are frozen to avoid overwriting during fine-tuning. A dynamic learning rate scheduler is used, applying a higher rate to the final task-specific layers and a lower rate to the frozen layers, aiding model adaptation while retaining pre-trained knowledge.

\section{Parameter-Efficient Modeling using Low-Rank Adaptation (LoRA)} \label{peft-lora}
\begin{figure}
    \centering
    \includegraphics[width=1.1\linewidth]{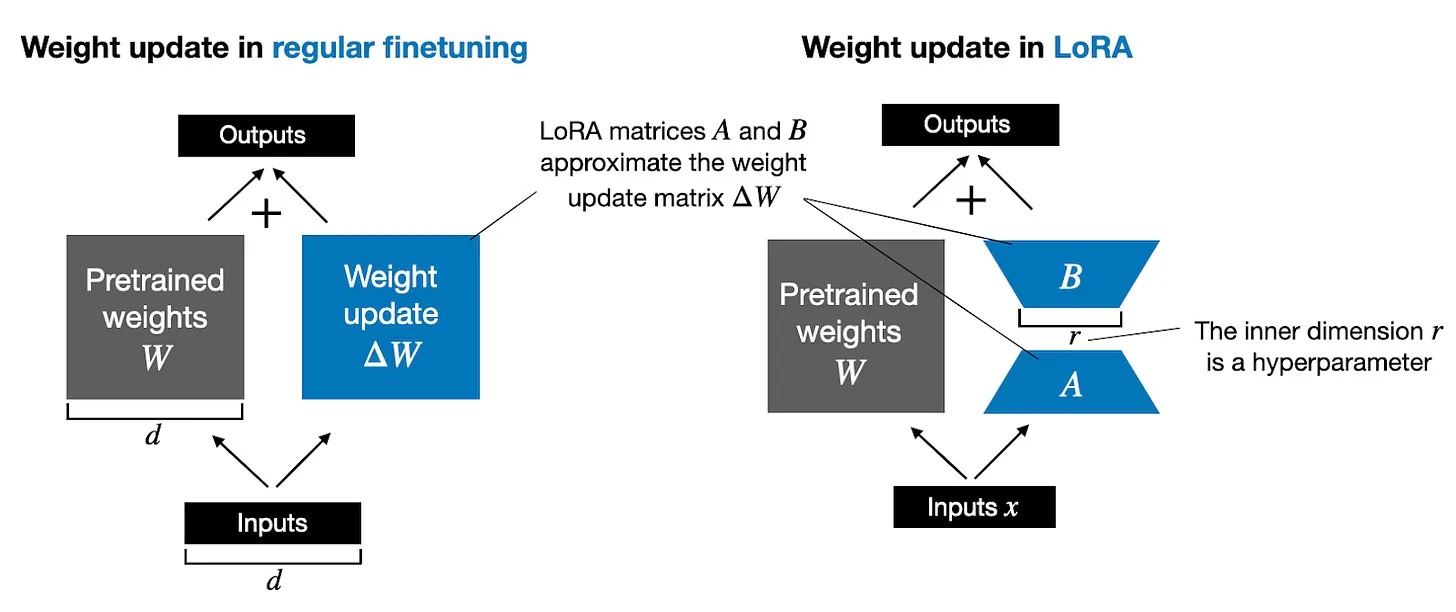}
    \caption{Weight update step \textit{(right}) with and \textit{(left)} without with LoRA adapter, Figure courtesy: \cite{sebraschka2024}}
    \label{fig:lora}
\end{figure}

\noindent We further explored Parameter Efficient Fine-Tuning  
 (\cite{houlsby2019parameterefficient}) for fine-tuning our pre-trained OPT 125M model on COLA dataset by employing the Low-Rank Adaptation (LoRA) adapter (\cite{hu2021lora}). In this method of fine-tuning a model, we break down the weight matrix that is learned during the gradient descent step of backpropagation into two, smaller-rank matrices, thereby significantly reducing the number of parameters that we optimize for during our training process. This is depicted in Figure \ref{fig:lora}. If say, in a regular training process, we learn a matrix $\Delta W$ of dimensions $m$ x $m$, such that:

\begin{equation}
W \xleftarrow{} W + \Delta W
\end{equation}

, with LoRA, we decompose the $\Delta W$ into smaller-rank matrices $A$ and $B$ of dimensions $m$ x $r$ and $r$ x $m$, such that: 

\begin{equation}
\begin{split}
\Delta W = A . B, \\
\ni r \ll m
\end{split}
\end{equation}

, where $r$, rank of the matrices, is a hyper-parameter of the model.

In the domain of LLMs, decomposing non-trivial weight matrices in each layer with r = 2, 4, 6, …, significantly reduces the parameters to optimize, depending on the chosen rank. This decomposition is an approximation that aims for a balance between model accuracy and resource utilization during fine-tuning. In LoRA, the adapter layer output is multiplied by a factor $\alpha$ (another hyper-paramter), which determines the impact of the LoRA decomposition on the model layer to which the adapter is applied.

\section{Context Distillation}
\noindent In our approach, training uses two primary loss functions: distillation and classification loss. Distillation loss, computed using Kullback-Leibler divergence (\cite{kldiv1975}), is a measure that quantifies the difference between any two probability distributions. More specifically, in our case, the KL divergence is calculated between the log-softmax of the student model's outputs and the probability distribution of the teacher model, with the results averaged across the batch. Classification loss, determined using cross-entropy loss function, assesses the discrepancy between the student model’s predictions and the true labels.

We integrate these losses into training by formulating the overall loss as a weighted sum of both, with each loss component assigned a task-dependent weight. This balance is represented by the equation:
\begin{equation}
\text{Loss} = 0.5 \times \text{Distillation Loss} + 0.5 \times \text{Classification Loss}
\end{equation}

This approach allows the student model to learn from hard labels and mimic the teacher model’s probabilistic output, fostering nuanced understanding and potential improved generalization on unseen data (\cite{askell2021general}).

We also assessed context distillation with few-shot learning setup from section \ref{Few-Shot Learning}, where the student model was fine-tuned using limited data. Subsets of representative samples were used to fine-tune the student model, and its performance was evaluated.

\section{Experiments and Results} \label{Experiments and Results}

\noindent This section examines the effectiveness of fine-tuning techniques on two OPT models: OPT 125M and OPT 350M. We cover Vanilla FT, PBFT, PEFT with LoRA adapters, Adaptive FT, and Context Distillation (with and without few-shot learning). These experiments helped us understand the model's adaptability to sequence classification and methods to improve its performance.

\subsection{Hyper-Parameters and Experiment Set-Up}
\noindent We initiated this empirical study by evaluating the impact of various hyper-parameters in our experiments. To compare the myriad techniques we developed against the benchmark fairly, we aligned our fine-tuning parameters with those in \cite{mosbach2023fewshot}. We used few-shot sample sizes of 2, 16, 32, 64, and 128 examples per class to reflect different levels of information availability. Our training spanned 40 epochs with a batch size of 32 to balance data exposure and computational efficiency. We conducted 10 trials per run to remove bias, selecting samples randomly from the training set. The learning rate was set at $1e^{-5}$, with zero weight decay to minimize over-penalization of model complexity. We set a warmup ratio at $10\%$ and employed the AdamW optimizer (\cite{loshchilov2019decoupled}), a typical choice for LLM optimization. A consolidated table of the hyper-parameters used in our experiments can be found in Appendix \ref{hyperparam}.

\subsection{Vanilla FT, PBFT and Adaptive FT}

\subsubsection{Vanilla FT} \label{vanilla-ft-sec}
\begin{figure}[h]
    \centering
    \begin{minipage}{0.5\columnwidth}
      \includegraphics[width=\linewidth]{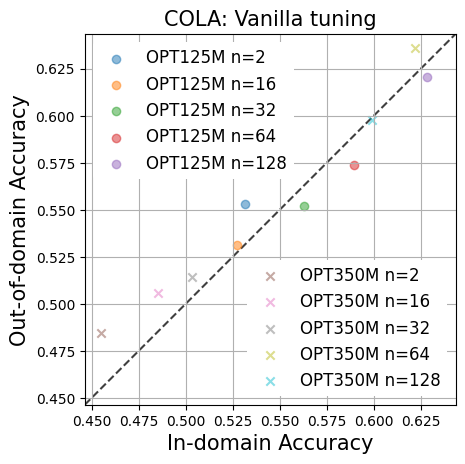}
    \end{minipage}%
    \begin{minipage}{0.5\columnwidth}
      \includegraphics[width=\linewidth]{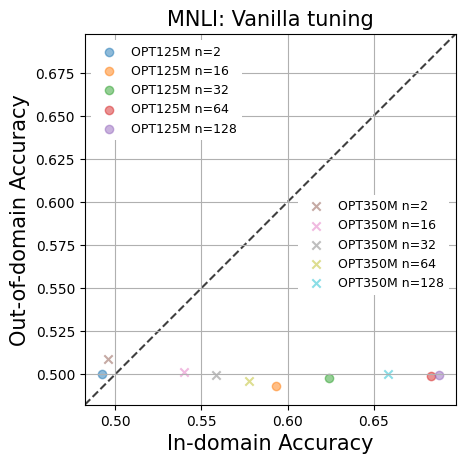}
    \end{minipage}
\caption{Vanilla FT accuracy for OPT 125M and OPT 350M on \textit{(left)} COLA and \textit{(right)} MNLI for different samples}
\label{fig:mnli-cola-vanilla}
\end{figure}

In our first experiment, we assessed the impact of vanilla fine-tuning on the CoLA and MNLI datasets using OPT 125M and OPT 350M models. The left plot in Figure \ref{fig:mnli-cola-vanilla} shows that for the CoLA dataset, both models exhibited increased accuracy with larger sample sizes, with the OPT 125M model peaking at N=128. OPT 350M model having nearly 3 times the parameters of the OPT 125M showed less pronounced effect of data size on accuracy.

 The right plot in Figure \ref{fig:mnli-cola-vanilla} presents a similar case for the MNLI data set. The only caveat was that models’ out-of-domain accuracy remained relatively stagnant across varying sample sizes. The larger OPT 350M model showed moderate enhancements in in-domain accuracy as the sample size grew, with minimal fluctuations in out-of-domain accuracy. This was expected because as detailed in Section \ref{Datasets}, MNLI dataset uses the HANS dataset as OOD data while the COLA dataset uses a subset of its own sources. The OPT 125M model consistently outperformed the OPT 350M model in in-domain accuracy across all few-shot learning sizes. This is due to the higher complexity of OPT 350M which did not allow better generalization. 

\subsubsection{PBFT} \label{pbft-sec}
\begin{figure}
    \begin{minipage}{0.5\columnwidth}
      \includegraphics[width=\linewidth]{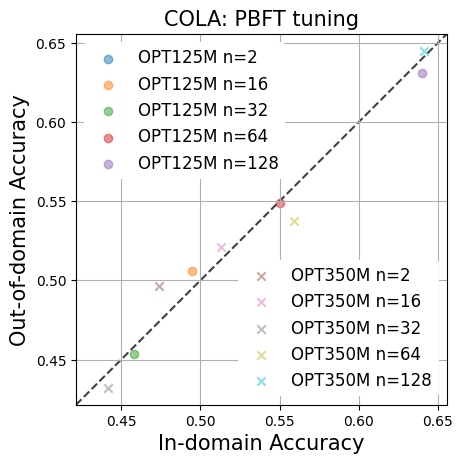}
    \end{minipage}%
    \begin{minipage}{0.5\columnwidth}
      \includegraphics[width=\linewidth]{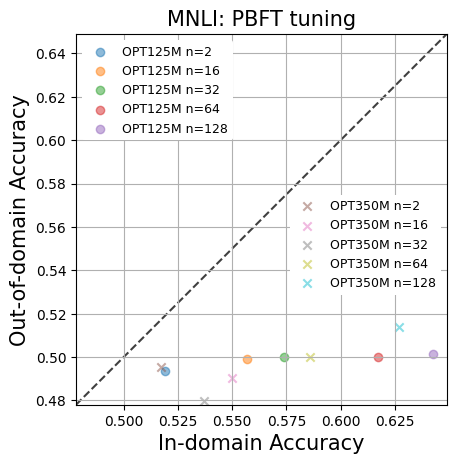}
    \end{minipage}
\caption{PBFT accuracy for OPT 125M and OPT 350M on \textit{(left)} COLA and \textit{(right)} MNLI for different samples}
\label{fig:mnli-cola-pbft}
\end{figure}

For the COLA dataset, as we can see in the left plot of Figure \ref{fig:mnli-cola-pbft}, OPT 125M model displayed better out-of-domain accuracy at the smallest sample size (N=2), while the OPT 350M model showed better in-domain performance at mid-range sample sizes (N=16, 64). At N=128, the OPT 350M model slightly surpassed the 125M model in out-of-domain accuracy, demonstrating its potential for higher performance gains at larger few-shot sizes. These findings highlight the complex effects of PBFT on model accuracy, suggesting larger few-shot sizes may be preferable for optimal performance in linguistic acceptability tasks. We used GPT-3 style prompts for our patterns, suggesting that improved prompting could further enhance our models’ performance. However, the study of prompts’ effect on model accuracy is beyond the scope of this work.

For the MNLI dataset, OPT 125M model consistently outperformed the OPT 350M model in in-domain accuracy at higher sample sizes, while out-of-domain accuracies remained relatively stable across both models, with slight improvements noted at the largest sample sizes. 

\subsubsection{Adaptive FT} \label{adaptive-ft-sec}
\begin{figure}[h]
    \centering
    \includegraphics[width=0.5\linewidth]{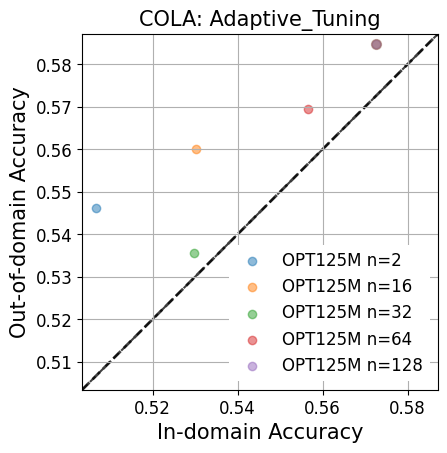}
    \caption{Adaptive FT accuracy for OPT 125M on COLA for different few-shot samples}
    \label{fig:adaptive-ft}
\end{figure}

We performed adaptive tuning on the OPT 125M model by freezing the entire model and tuning only the last two decoder layers (11th and 12th). We used a dynamic learning rate strategy. This approach aimed to assess the impact of targeted layer freezing and responsive learning rate adjustments on model performance. Our results showed progressive improvement in both in-domain and out-of-domain accuracies as sample size increased, peaking at N=128 as seen in Figure \ref{fig:adaptive-ft}. These findings underscore the effectiveness of this approach in boosting performance and enhancing generalization capabilities. However, compared to full-model fine-tuning, \ref{vanilla-ft-sec}  and \ref{pbft-sec}, the accuracies obtained in adaptive fine-tuning were lower, as expected, since we were only tuning the last few layers of the model.

\subsection{Effect of using parameter-efficient modeling} \label{exp:peft-lora}

\noindent We anticipated similar performance between parameter-efficient fine-tuning (PEFT), vanilla, and pattern-based fine-tuning. Experiments were conducted for different matrix ranks, adapting all model layers with LoRA adapters. The LoRA configuration was applied to training but not inference. After experimenting with different $\alpha$ and LoRA dropout values, we settled on 8 and 0.1 respectively.

\begin{figure}[h]
    \centering
    \includegraphics[width=0.5\linewidth]{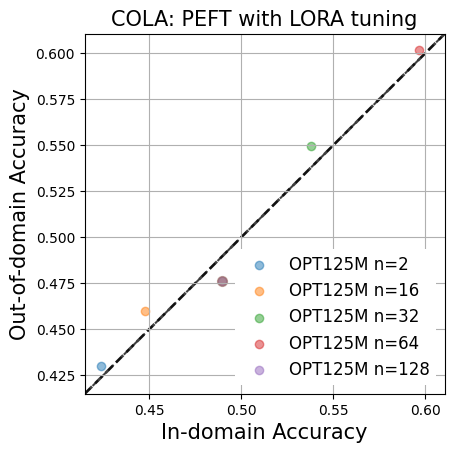}
    \caption{Accuracy on COLA for various few shot samples averaged across ranks of LoRA layers}
    \label{fig:lora-cola}
\end{figure}

In Figure \ref{fig:lora-cola} -- The plot compares in-domain and out-of-domain accuracies on the COLA dataset for different example sizes, showing an upward trend for N=2 to N=64, highlighting the benefits of more training data. However, accuracy decreases at N=128, a trend differing from other tuning methods used in this study. This suggests that the matrix decomposition approximation might impact the adaptability of the original model for the inferred task. Given the OPT model’s design for sequence generation and its use for classification here, techniques like LoRA and few-shot, which add approximation to backpropagation and reduce the training set, might not be as effective. We refer user to Appendix \ref{all-lora} for the graphs for individual ranks. 

These results illuminate the nuanced balance required in parameter-efficient fine-tuning strategies to enhance model performance while maintaining robust generalization capabilities.

\subsection{Effect of using context distillation}

\noindent We expected two outcomes from context distillation. Firstly, the distilled models would likely show better performance due to acquired knowledge. Secondly, the few-shot distillation aimed to balance performance with efficiency, enhancing learning with fewer examples, potentially leading to more resource-efficient training without much performance compromise.

Applying context distillation to the OPT 125M model showed a domain accuracy increase from 0.6314 to 0.7209 over three epochs, confirming its effectiveness. However, out-of-domain accuracy was inconsistent, peaking at 0.5250 then dropping to 0.4975, showing complex interactions between domain-specific learning and generalization.

 \begin{figure}
    \begin{minipage}{0.5\columnwidth}
      \includegraphics[width=\linewidth]{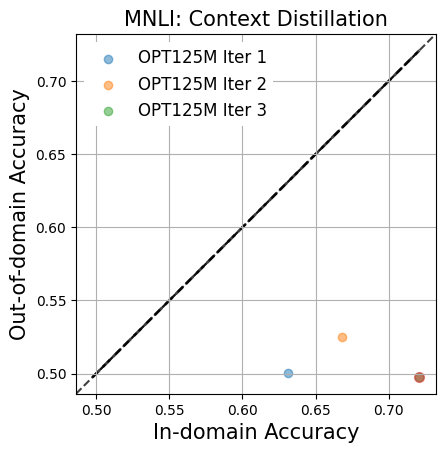}
    \end{minipage}%
    \begin{minipage}{0.5\columnwidth}
      \includegraphics[width=\linewidth]{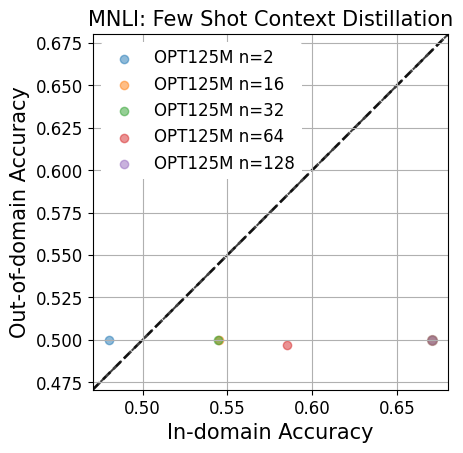}
    \end{minipage}
\caption{Accuracy with context distillation on OPT 125M model \textit{(right)} with and \textit{(left)} w/o few-shot on MNLI for different samples}
\label{fig:mnli-context distillation}
\end{figure}

Few-shot distillation evaluations showed a significant increase in in-domain accuracy from 0.4801 at N=2 to 0.6708 at N=128, indicating better knowledge assimilation with more examples. Out-of-domain accuracy remained stable at 0.5, with a slight drop to 0.4967 at N=64, suggesting limited generalization benefits. Combining few-shot with context distillation achieved an in-domain accuracy of 0.67 with N=128, nearing the 0.72 from traditional context distillation. This combined approach took only 5 minutes and 19 seconds, compared to 1 hour and 51 minutes for traditional methods, suggesting that increasing sample size in few-shot fine-tuning could match or surpass full dataset fine-tuning efficacy, while being more time and resource-efficient.

Additionally, it is important to note that no significant performance differences were observed in the out-of-domain evaluation, using the HANS dataset.

\section{Conclusion} \label{conclusion}

\noindent In this paper, we compared different ways of efficiently fine-tuning a LLM originally created for sequence generation tasks on a sequence classification task in the realms of few-shot learning. We examined standard fine-tuning techniques like vanilla FT, PBFT and compared their OOD accuracies against some of the more advanced methods such as adaptive FT, PEFT with LoRA and context distillation. We present results for a few-shot setting with 16 examples ($N$) in this section, with additional comparisons in Appendix \ref{all-few-shot}.

As we can see in Table \ref{tab:vanilla-vs-pbft} below, Our experiments reveal that vanilla FT outperforms PBFT on the CoLA dataset for the smaller model, indicating a need for improved prompting. 

\begin{table}[ht]
\centering
\caption{OOD accuracy comparison b/w Vanilla and PBFT on COLA (N=16)}
\vspace{5pt}
\label{tab:vanilla-vs-pbft}
\begin{tabular}{lcc}
\toprule
\midrule
Model & OPT 125M & OPT 350M \\
Method &  &  \\
\midrule
\textbf{Vanilla Fine Tuning} & 0.5310 & 0.5058 \\
\textbf{Pattern Based} & 0.5058 & 0.5213 \\
\bottomrule
\end{tabular}
\end{table}

Adaptive fine-tuning performs better compared to vanilla fine-tuning and pattern-based methods (Table \ref{tab:adpft-vs-others}), suggesting its effectiveness in optimizing the model for better generalization in certain contexts.

\begin{table}[h]
\centering
\caption{OOD delta b/w adaptive and base FT on COLA (N=16)}
\vspace{5pt}
\label{tab:adpft-vs-others}
\begin{tabular}{l|c} 
\textbf{Method} & \textbf{Adaptive Fine Tuning} \\ \hline

\midrule
\textbf{Vanilla Fine Tuning} & 0.0290 \\
\textbf{Pattern Based} & 0.0543 \\
\bottomrule
\end{tabular}
\end{table}

Accuracies in LoRA are similar to other fine-tuning methods  (Table \ref{peft-lora-table}), potentially due to fine-tuning all layers in the latter set of methods. Also, as noted earlier, we use a $\alpha$ value of 8 for our LoRA configuration, as compared to 32 by the original authors of the paper \cite{hu2021lora}, which could undermine LoRA.

\begin{table}[ht]
\centering
\caption{OOD delta b/w LoRA and base FT on COLA (N=16)}
\label{peft-lora-table}
{\small 
\begin{tabular}{@{}lrrrrr@{}}
\toprule
\textbf{Method/Rank} & \multicolumn{1}{l}{1} & \multicolumn{1}{l}{2} & \multicolumn{1}{l}{4} & \multicolumn{1}{l}{8} & \multicolumn{1}{l}{64} \\ \midrule
\textbf{Vanilla FT} & -0.0755 & -0.0697 & -0.0697 & -0.0697 & -0.0697 \\
\textbf{Pattern Based}       & -0.0503 & -0.0445 & -0.0445 & -0.0445 & -0.0445 \\ \bottomrule
\end{tabular}
} 
\end{table}

Finally, we notice that context distillation leads to a modest improvement in OOD accuracy over the other methods (Table \ref{tab:cd-vs-others}). Despite small differences, it underscores context distillation’s potential to boost the model’s generalization beyond pattern-based or vanilla fine-tuning techniques, explaining its growing popularity in the LLM community.

\begin{table}[!h]
\centering
\caption{OOD delta b/w few shot context distillation and base FT on MNLI (N=16)}
\vspace{5pt}
\label{tab:cd-vs-others}
\begin{tabular}{l|c} 
\textbf{Method} & \textbf{Few Shot Context Distillation} \\ \hline

\midrule
\textbf{Vanilla Fine Tuning} & 0.0310 \\
 \textbf{Pattern Based } & 0.0058 \\
\bottomrule
\end{tabular}
\end{table}

In summary, our analysis highlights the advantages and subtleties of various fine-tuning methods in the LLM community. We consolidated these methods in a few-shot learning setting to understand their benefits. Despite resource limitations, our experiments establish a premise and more extensive resources and time could yield a deeper understanding of these methods.

\section{Acknowledgement} \label{ack}

We would like to thank Dr. Zsolt Kira of Georgia Institute of Technology and the course teaching assistants and staff of CS 7643: Deep Learning at Georgia Institute of Technology. This work was pursued as part of the above-said course and the knowledge gained by the authors in this course was instrumental in carrying out this work.

\bibliographystyle{unsrtnat}
\bibliography{references}  

\begin{thebibliography}{14}
\providecommand{\natexlab}[1]{#1}
\providecommand{\url}[1]{\texttt{#1}}
\expandafter\ifx\csname urlstyle\endcsname\relax
  \providecommand{\doi}[1]{doi: #1}\else
  \providecommand{\doi}{doi: \begingroup \urlstyle{rm}\Url}\fi

\bibitem[Mosbach et~al.(2023)Mosbach, Pimentel, Ravfogel, Klakow, and
  Elazar]{mosbach2023fewshot}
Marius Mosbach, Tiago Pimentel, Shauli Ravfogel, Dietrich Klakow, and Yanai
  Elazar.
\newblock Few-shot fine-tuning vs. in-context learning: A fair comparison and
  evaluation, 2023.

\bibitem[Hu et~al.(2021)Hu, Shen, Wallis, Allen-Zhu, Li, Wang, Wang, and
  Chen]{hu2021lora}
Edward~J. Hu, Yelong Shen, Phillip Wallis, Zeyuan Allen-Zhu, Yuanzhi Li, Shean
  Wang, Lu~Wang, and Weizhu Chen.
\newblock Lora: Low-rank adaptation of large language models, 2021.

\bibitem[Williams et~al.(2018)Williams, Nangia, and Bowman]{N18-1101}
Adina Williams, Nikita Nangia, and Samuel Bowman.
\newblock A broad-coverage challenge corpus for sentence understanding through
  inference.
\newblock In \emph{Proceedings of the 2018 Conference of the North American
  Chapter of the Association for Computational Linguistics: Human Language
  Technologies, Volume 1 (Long Papers)}, pages 1112--1122. Association for
  Computational Linguistics, 2018.
\newblock URL \url{http://aclweb.org/anthology/N18-1101}.

\bibitem[Warstadt et~al.(2018)Warstadt, Singh, and Bowman]{warstadt2018neural}
Alex Warstadt, Amanpreet Singh, and Samuel~R Bowman.
\newblock Neural network acceptability judgments.
\newblock \emph{arXiv preprint arXiv:1805.12471}, 2018.

\bibitem[Zhang et~al.(2022)Zhang, Roller, Goyal, Artetxe, Chen, Chen, Dewan,
  Diab, Li, Lin, Mihaylov, Ott, Shleifer, Shuster, Simig, Koura, Sridhar, Wang,
  and Zettlemoyer]{zhang2022opt}
Susan Zhang, Stephen Roller, Naman Goyal, Mikel Artetxe, Moya Chen, Shuohui
  Chen, Christopher Dewan, Mona Diab, Xian Li, Xi~Victoria Lin, Todor Mihaylov,
  Myle Ott, Sam Shleifer, Kurt Shuster, Daniel Simig, Punit~Singh Koura, Anjali
  Sridhar, Tianlu Wang, and Luke Zettlemoyer.
\newblock Opt: Open pre-trained transformer language models, 2022.

\bibitem[Brown et~al.(2020)Brown, Mann, Ryder, Subbiah, Kaplan, Dhariwal,
  Neelakantan, Shyam, Sastry, Askell, Agarwal, Herbert-Voss, Krueger, Henighan,
  Child, Ramesh, Ziegler, Wu, Winter, Hesse, Chen, Sigler, Litwin, Gray, Chess,
  Clark, Berner, McCandlish, Radford, Sutskever, and Amodei]{brown2020language}
Tom~B. Brown, Benjamin Mann, Nick Ryder, Melanie Subbiah, Jared Kaplan,
  Prafulla Dhariwal, Arvind Neelakantan, Pranav Shyam, Girish Sastry, Amanda
  Askell, Sandhini Agarwal, Ariel Herbert-Voss, Gretchen Krueger, Tom Henighan,
  Rewon Child, Aditya Ramesh, Daniel~M. Ziegler, Jeffrey Wu, Clemens Winter,
  Christopher Hesse, Mark Chen, Eric Sigler, Mateusz Litwin, Scott Gray,
  Benjamin Chess, Jack Clark, Christopher Berner, Sam McCandlish, Alec Radford,
  Ilya Sutskever, and Dario Amodei.
\newblock Language models are few-shot learners, 2020.

\bibitem[Wang et~al.(2019)Wang, Singh, Michael, Hill, Levy, and
  Bowman]{wang2019glue}
Alex Wang, Amanpreet Singh, Julian Michael, Felix Hill, Omer Levy, and
  Samuel~R. Bowman.
\newblock Glue: A multi-task benchmark and analysis platform for natural
  language understanding, 2019.

\bibitem[McCoy et~al.(2019)McCoy, Pavlick, and Linzen]{mccoy2019right}
Tom McCoy, Ellie Pavlick, and Tal Linzen.
\newblock Right for the wrong reasons: Diagnosing syntactic heuristics in
  natural language inference.
\newblock In \emph{Proceedings of the 57th Annual Meeting of the Association
  for Computational Linguistics}, pages 3428--3448, Florence, Italy, 2019.
  Association for Computational Linguistics.

\bibitem[Inc.(2023)]{meta2023efficientllm}
Meta~Platforms Inc.
\newblock Few-shot fine-tuning vs. in-context learning: A fair comparison and
  evaluation.
\newblock \emph{https://github.com/uds-lsv/llmft}, 2023.

\bibitem[Askell et~al.(2021)Askell, Bai, Chen, Drain, Ganguli, Henighan, Jones,
  Joseph, Mann, DasSarma, Elhage, Hatfield-Dodds, Hernandez, Kernion, Ndousse,
  Olsson, Amodei, Brown, Clark, McCandlish, Olah, and
  Kaplan]{askell2021general}
Amanda Askell, Yuntao Bai, Anna Chen, Dawn Drain, Deep Ganguli, Tom Henighan,
  Andy Jones, Nicholas Joseph, Ben Mann, Nova DasSarma, Nelson Elhage, Zac
  Hatfield-Dodds, Danny Hernandez, Jackson Kernion, Kamal Ndousse, Catherine
  Olsson, Dario Amodei, Tom Brown, Jack Clark, Sam McCandlish, Chris Olah, and
  Jared Kaplan.
\newblock A general language assistant as a laboratory for alignment, 2021.

\bibitem[Houlsby et~al.(2019)Houlsby, Giurgiu, Jastrzebski, Morrone,
  de~Laroussilhe, Gesmundo, Attariyan, and
  Gelly]{houlsby2019parameterefficient}
Neil Houlsby, Andrei Giurgiu, Stanislaw Jastrzebski, Bruna Morrone, Quentin
  de~Laroussilhe, Andrea Gesmundo, Mona Attariyan, and Sylvain Gelly.
\newblock Parameter-efficient transfer learning for nlp, 2019.

\bibitem[Raschka(2024)]{sebraschka2024}
Sebastian Raschka.
\newblock Improving lora: Implementing weight-decomposed low-rank adaptation
  (dora) from scratch.
\newblock
  \emph{https://magazine.sebastianraschka.com/p/lora-and-dora-from-scratch},
  2024.

\bibitem[Csiszar(1975)]{kldiv1975}
I.~Csiszar.
\newblock $i$-divergence geometry of probability distributions and minimization
  problems.
\newblock In \emph{The Annals of Probability}, pages 146 -- 158. The Annals of
  Probability, 1975.
\newblock URL \url{https://doi.org/10.1214/aop/1176996454}.

\bibitem[Loshchilov and Hutter(2019)]{loshchilov2019decoupled}
Ilya Loshchilov and Frank Hutter.
\newblock Decoupled weight decay regularization, 2019.

\end{thebibliography}

\newpage
\pagenumbering{Roman}

\appendix

\section{Appendix: Hyper-parameters} \label{hyperparam}

To perform a fair comparison between all the experiments, we kept the hyper-parameters uniform across all of them, as listed in table 
\ref{tab:all-params}.

\begin{table}[!h]
\centering
\caption{Hyper-parameters used in all the experiments}
\vspace{5pt}
\label{tab:all-params}
\begin{tabular}{l|c} 
\textbf{Hyper-parameter} & \textbf{Value} \\ \hline

\midrule
\textbf{Few-shot sample sizes} & {2, 16, 32, 64, 128} \\
\textbf{\# of epochs} & 40 \\
\textbf{Batch size} & 32 \\
\textbf{Learning Rate} & $1e^{-5}$ \\
\textbf{Weight decay} & 0.0 \\
\textbf{Warm-up ratio} & 0.1 \\
\textbf{\# of runs/trials per sample size} & 10 \\
\textbf{Optimizer} & AdamW \\
\bottomrule
\end{tabular}
\end{table}

Further, some additional parameters were used for experiments that involved LoRA. Those are listed below in Table \ref{tab:lora-params}.

\begin{table}[!h]
\centering
\caption{Additional hyper-parameters used in LoRA experiments}
\vspace{5pt}
\label{tab:lora-params}
\begin{tabular}{l|c} 
\textbf{Hyper-parameter} & \textbf{Value} \\ \hline

\midrule
\textbf{LoRA matrix ranks} & {1, 2, 4, 8, 64} \\
\textbf{$\alpha$} & 8 \\
\textbf{Dropout} & 0.1 \\
\bottomrule
\end{tabular}
\end{table}

\newpage

\section{Appendix: Plots for LoRA for different ranks} \label{all-lora}

In Section \ref{exp:peft-lora}, we plotted the results of our experiments with PEFT LoRA adapters for different few-shot sample sizes, averaged across ranks. In this appendix section, we present the supplementary plots for the same experiment but segregated for each individual matrix rank.

\begin{figure}[ht]
    \centering
    \includegraphics[width=0.4\columnwidth]{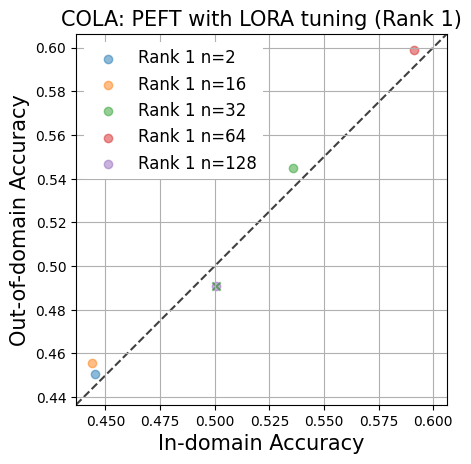}
    \caption{Accuracy for various few shot samples LoRA layers with rank, $r=1$}
    \label{fig:enter-label}
\end{figure}

\begin{figure}[ht]
    \centering
    \includegraphics[width=0.4\columnwidth]{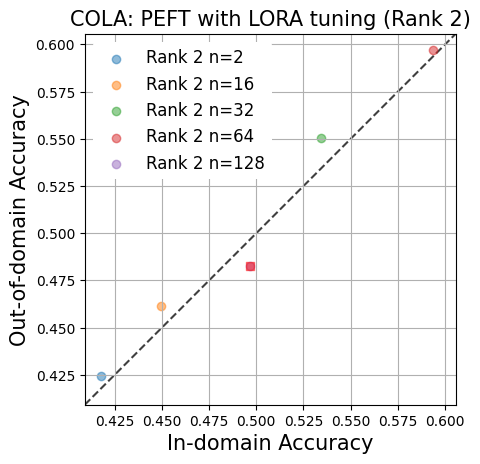}
    \caption{Accuracy for various few shot samples LoRA layers with rank, $r=2$}
    \label{fig:enter-label}
\end{figure}

\begin{figure}[ht]
    \centering
    \includegraphics[width=0.4\columnwidth]{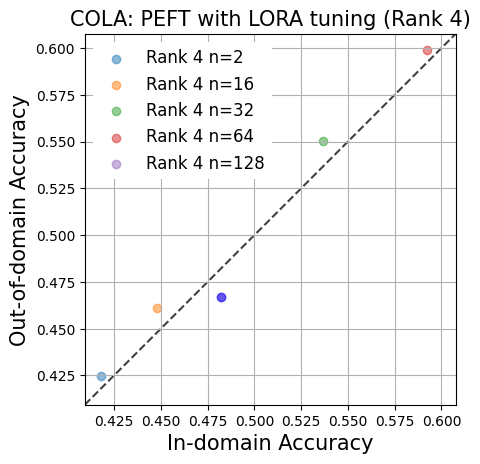}
    \caption{Accuracy for various few shot samples LoRA layers with rank, $r=4$}
    \label{fig:enter-label}
\end{figure}

\begin{figure}[ht]
    \centering
    \includegraphics[width=0.4\columnwidth]{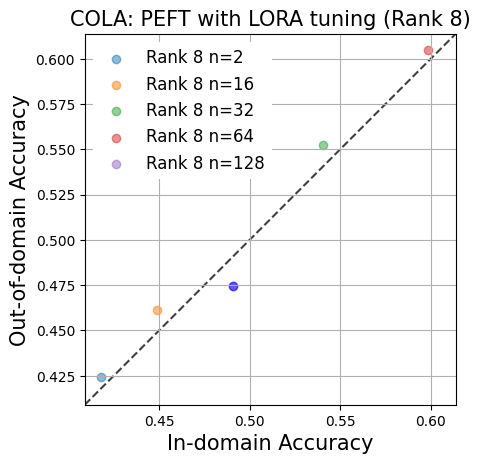}
    \caption{Accuracy for various few shot samples LoRA layers with rank, $r=8$}
    \label{fig:enter-label}
\end{figure}

\clearpage

\begin{figure}[H]
    \centering
    \includegraphics[width=0.4\columnwidth]{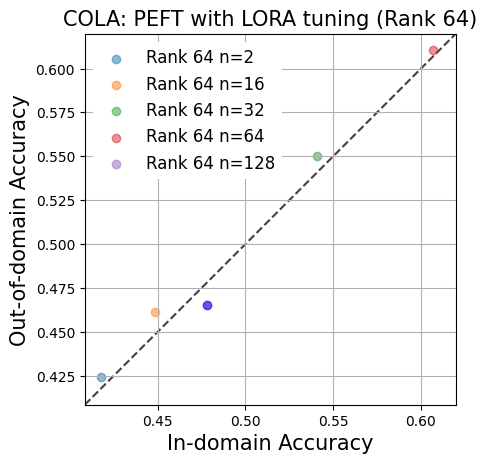}
    \caption{Accuracy for various few shot samples LoRA layers with rank, $r=64$}
    \label{fig:enter-label}
\end{figure}

\clearpage
\newpage
\section{Appendix: Results for various few-shot sample sizes} \label{all-few-shot}

In our conclusion section, Section \ref{conclusion}, for concise analysis and due to space constraint, we presented all the comparisons with respect to few-shot sample size, $N=16$. In this appndix section, we tabulate those comparisons for all the fine-tuning methods for all different few-shot sample sizes.

\subsection{Adaptive Fine-Tuning}

\begin{table}[h]
\centering
\caption{OOD delta b/w adaptive and base FT on COLA (N=2)}
\vspace{5pt}
\label{tab:adpft-vs-others-appdx}
\begin{tabular}{l|c} 
\textbf{Method} & \textbf{Adaptive Fine Tuning} \\ \hline

\midrule
\textbf{Vanilla Fine Tuning} & -0.006977
 \\
\textbf{Pattern Based} & -0.02441 \\
\bottomrule
\end{tabular}
\end{table}

\begin{table}[h]
\centering
\caption{OOD delta b/w adaptive and base FT on COLA (N=32)}
\vspace{5pt}
\label{tab:adpft-vs-others-appdx}
\begin{tabular}{l|c} 
\textbf{Method} & \textbf{Adaptive Fine Tuning} \\ \hline

\midrule
\textbf{Vanilla Fine Tuning} & -0.01647
 \\
\textbf{Pattern Based} & 0.0823 \\
\bottomrule
\end{tabular}
\end{table}

\begin{table}[!h]
\centering
\caption{OOD delta b/w adaptive and base FT on COLA (N=64)}
\vspace{5pt}
\label{tab:adpft-vs-others-appdx}
\begin{tabular}{l|c} 
\textbf{Method} & \textbf{Adaptive Fine Tuning} \\ \hline

\midrule
\textbf{Vanilla Fine Tuning} & -0.0040
 \\
\textbf{Pattern Based} & 0.0209 \\
\bottomrule
\end{tabular}
\end{table}

\begin{table}[!h]
\centering
\caption{OOD delta b/w adaptive and base FT on COLA (N=128)}
\vspace{5pt}
\label{tab:adpft-vs-others-appdx}
\begin{tabular}{l|c} 
\textbf{Method} & \textbf{Adaptive Fine Tuning} \\ \hline

\midrule
\textbf{Vanilla Fine Tuning} & -0.0358
 \\
\textbf{Pattern Based} & -0.04651 \\
\bottomrule
\end{tabular}
\end{table}


\subsection{Parameter-Efficient Fine tuning with Low-Rank Adaptation (LoRA)}

\begin{table}[!ht]
\centering
\caption{OOD delta b/w LoRA and base FT on COLA (N=2)}
\label{peft-lora-table-appdx}
{\small 
\begin{tabular}{@{}lrrrrr@{}}
\toprule
Method/Rank & \multicolumn{1}{l}{1} & \multicolumn{1}{l}{2} & \multicolumn{1}{l}{4} & \multicolumn{1}{l}{8} & \multicolumn{1}{l}{64} \\ \midrule
Vanilla FT & -0.1025 & -0.1286 & -0.1286 & -0.1286 & -0.1286 \\
Pattern Based       & -0.1199 & -0.1461 & -0.1461 & -0.1461 & -0.1461 \\ \bottomrule
\end{tabular}
} 
\end{table}

\begin{table}[!ht]
\centering
\caption{OOD delta b/w LoRA and base FT on COLA (N=32)}
\label{peft-lora-table-appdx}
{\small 
\begin{tabular}{@{}lrrrrr@{}}
\toprule
Method/Rank & \multicolumn{1}{l}{1} & \multicolumn{1}{l}{2} & \multicolumn{1}{l}{4} & \multicolumn{1}{l}{8} & \multicolumn{1}{l}{64} \\ \midrule
Vanilla FT & -0.0073 & -0.0017 & -0.0017 & 0.0001 & -0.0017 \\
Pattern Based       & 0.0914 & 0.0970 & 0.0970 & 0.0990 & 0.0970 \\ \bottomrule
\end{tabular}
} 
\end{table}

\begin{table}[!ht]
\centering
\caption{OOD delta b/w LoRA and base FT on COLA (N=64)}
\label{peft-lora-table-appdx}
{\small 
\begin{tabular}{@{}lrrrrr@{}}
\toprule
Method/Rank & \multicolumn{1}{l}{1} & \multicolumn{1}{l}{2} & \multicolumn{1}{l}{4} & \multicolumn{1}{l}{8} & \multicolumn{1}{l}{64} \\ \midrule
Vanilla FT & 0.0251 & 0.0232 & 0.0251 & 0.0310 & 0.0368 \\
Pattern Based       & 0.0501 & 0.0482 & 0.0501 & 0.0560 & 0.0618 \\ \bottomrule
\end{tabular}
} 
\end{table}

\begin{table}[!ht]
\centering
\caption{OOD delta b/w LoRA and base FT on COLA (N=128)}
\label{peft-lora-table-appdx}
{\small 
\begin{tabular}{@{}lrrrrr@{}}
\toprule
Method/Rank & \multicolumn{1}{l}{1} & \multicolumn{1}{l}{2} & \multicolumn{1}{l}{4} & \multicolumn{1}{l}{8} & \multicolumn{1}{l}{64} \\ \midrule
Vanilla FT & -0.1296 & -0.1379 & -0.1534 & -0.1457 & -0554 \\
Pattern Based       & -0.1403 & -0.1486 & -0.1641 &-0.1563 & -0.1660 \\ \bottomrule
\end{tabular}
} 
\end{table}

\subsection{Context Distillation with Few-Shot Learning}

\begin{table}[!h]
\centering
    \caption{OOD delta b/w context distillation and base FT on MNLI (N=2)}
    \vspace{5pt}
    \label{tab:cd-vs-others-appdx}
    \begin{tabular}{l|c} 
    \textbf{Method} & \textbf{Few Shot Context Distillation} \\ \hline
    
    \midrule
    \textbf{Vanilla Fine Tuning} & -0.0531 \\
    \textbf{Pattern Based} & -0.07054 \\
    \bottomrule
    \end{tabular}
\end{table}

\begin{table}[!h]
\centering
    \caption{OOD delta b/w context distillation and base FT on MNLI (N=32)}
    \vspace{5pt}
    \label{tab:cd-vs-others-appdx}
    \begin{tabular}{l|c} 
    \textbf{Method} & \textbf{Few Shot Context Distillation} \\ \hline
    
    \midrule
    \textbf{Vanilla Fine Tuning} & 0.0521 \\
    \textbf{Pattern Based} & 0.0467 \\
    \bottomrule
    \end{tabular}
\end{table}

\begin{table}[!h]
\centering
    \caption{OOD delta b/w context distillation and base FT on MNLI (N=64)}
    \vspace{5pt}
    \label{tab:cd-vs-others-appdx}
    \begin{tabular}{l|c} 
    \textbf{Method} & \textbf{Few Shot Context Distillation} \\ \hline
    
    \midrule
    \textbf{Vanilla Fine Tuning} & 0.0769 \\
    \textbf{Pattern Based} & 0.05191 \\
    \bottomrule
    \end{tabular}
\end{table}

\begin{table}[!h]
\centering
    \caption{OOD delta b/w context distillation and base FT on MNLI (N=128)}
    \vspace{5pt}
    \label{tab:cd-vs-others-appdx}
    \begin{tabular}{l|c} 
    \textbf{Method} & \textbf{Few Shot Context Distillation} \\ \hline
    
    \midrule
    \textbf{Vanilla Fine Tuning} & 0.1205 \\
    \textbf{Pattern Based} & 0.1312 \\
    \bottomrule
    \end{tabular}
\end{table}

\end{document}